\documentclass{article}

\usepackage[preprint,nonatbib]{neurips_2022}

\usepackage[preprint]{neurips_2022}

\usepackage[utf8]{inputenc} % allow utf-8 input
\usepackage[T1]{fontenc}    % use 8-bit T1 fonts
\usepackage{hyperref}       % hyperlinks
\usepackage{url}            % simple URL typesetting
\usepackage{booktabs}       % professional-quality tables
\usepackage{amsfonts}       % blackboard math symbols
\usepackage{nicefrac}       % compact symbols for 1/2, etc.
\usepackage{microtype}      % microtypography
\usepackage{xcolor}         % colors

\usepackage[
    natbib=true,
    citestyle=numeric-comp,
    bibstyle=ieee,
    sorting=nty
]{biblatex}
\usepackage{graphicx}
\usepackage{caption}
\usepackage{subcaption}
\usepackage{bm}
\usepackage{amsmath,amssymb}
\usepackage{booktabs}
\usepackage{algorithm}
\usepackage[noend]{algorithmic}
\usepackage{multirow}
\usepackage[flushleft]{threeparttable}
\usepackage{makecell}

\hypersetup{
    colorlinks=true,
    linkcolor=red,
    citecolor=cyan,
    filecolor=magenta,      
    urlcolor=cyan,
}

\definecolor{f3green}{RGB}{0,144,81}
\definecolor{f3brown}{RGB}{148,82,0}

\usepackage[textsize=tiny]{todonotes}
\usepackage{hhline}
\usepackage{colortbl}
\definecolor{Gray}{gray}{0.9}
\definecolor{ForestGreen}{rgb}{0.13, 0.55, 0.13}
\definecolor{ForestGreen2}{rgb}{0.2, 0.5372549019607843, 0.1803921568627451}
\definecolor{newGREEN}{RGB}{34,139,34}
\definecolor{myMaroon}{RGB}{231, 52, 52}
\definecolor{Maroon}{rgb}{0.69, 0.19, 0.0}
\definecolor{my_cyan}{RGB}{112, 242, 244}

\usepackage{wrapfig}

\title{
    \makebox[\textwidth][c]{
        \makebox[2.5em][l]{\raisebox{-0.4\height}{\includegraphics[height=2em]{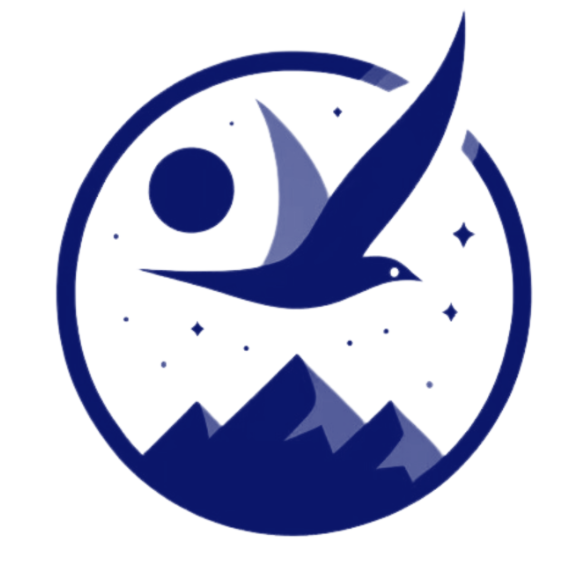}}}
        \hspace{-2.5em} % 图标和文字之间的间距
        \parbox[c]{0.8\textwidth}{ % 文字框，宽度设置为整行的 60%
            \centering % 文字在框内居中
            Open-Sora: Democratizing Efficient\\Video Production for All
        }
    }
}

\author{
Zangwei Zheng$^*$, \And Xiangyu Peng\footnotemark, \And
Tianji Yang, \And Chenhui Shen, \And
Shenggui Li, \And
Hongxin Liu, \And Yukun Zhou, \And Tianyi Li, \And Yang You \AND
HPC-AI Tech
}

\begin{document}

\maketitle

\begin{abstract}
% pxy
Vision and language are the two foundational senses for humans, and they build up our cognitive ability and intelligence. While significant breakthroughs have been made in AI language ability, artificial visual intelligence, especially the ability to generate and simulate the world we see, is far lagging behind. To facilitate the development and accessibility of artificial visual intelligence, we created Open-Sora, an open-source video generation model designed to produce high-fidelity video content. Open-Sora supports a wide spectrum of visual generation tasks, including text-to-image generation, text-to-video generation, and image-to-video generation. The model leverages advanced deep learning architectures and training/inference techniques to enable flexible video synthesis, which could generate video content of up to 15 seconds, up to 720p resolution, and arbitrary aspect ratios. Specifically, we introduce Spatial-Temporal Diffusion Transformer (STDiT), an efficient diffusion framework for videos that decouples spatial and temporal attention. We also introduce a highly compressive 3D autoencoder to make representations compact and further accelerate training with an ad hoc training strategy. Through this initiative, we aim to foster innovation, creativity, and inclusivity within the community of AI content creation. By embracing the open-source principle, Open-Sora democratizes full access to all the training/inference/data preparation codes as well as model weights. All resources are publicly available at: \url{https://github.com/hpcaitech/Open-Sora}.

\begin{figure}[!h]
\begin{center}
\centerline{\includegraphics[width=0.84\textwidth]{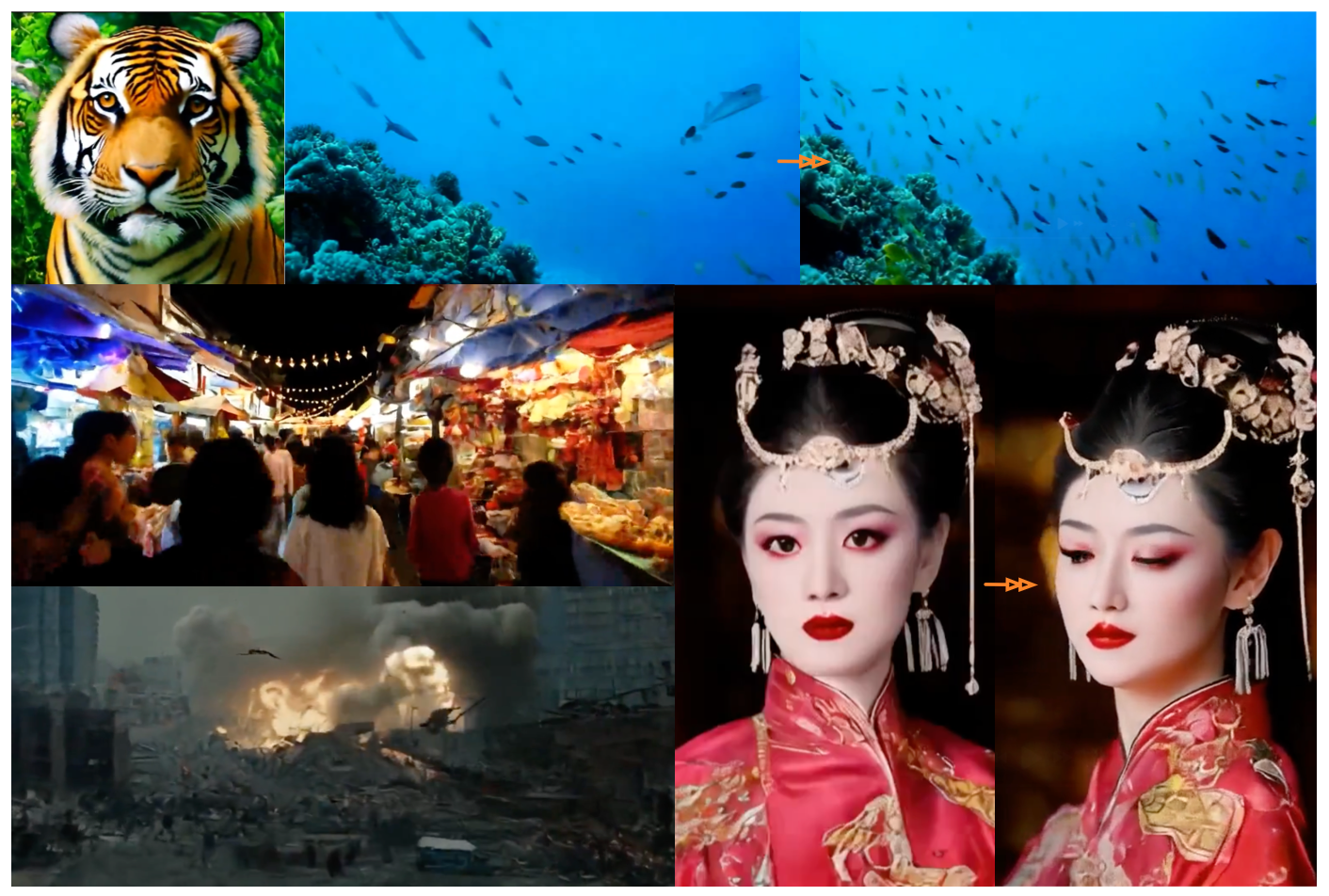}}
\caption{Open-Sora can generate high-fidelity videos. Images with arrows illustrate the motion.}
\label{fig:header}
\end{center}
\end{figure}

% Video generation has recently gained significant attention, particularly following the release of OpenAI Sora. However, the proprietary nature of Sora has limited its accessibility. In response, we introduce Open-Sora, an open-source video generation model designed to enable high-quality and flexible video synthesis. Open-Sora supports a range of tasks, including text-to-image, text-to-video, and image-to-video generation, with output durations up to 15 seconds, resolutions up to 720p, and arbitrary aspect ratios. Our implementation replicates key innovations from the Sora framework, such as arbitrary resolution training and a 3D autoencoder, while providing comprehensive access to model weights, data processing pipelines, and training code. We also introduce spatial-temporal attention and an efficient training strategy to reduce the training cost. By adhering to open-source principles, Open-Sora democratizes access to state-of-the-art video generation technologies, fostering innovation and collaboration within the research community. Open-Sora is currently the state-of-the-art 1B-scale model in the open-source video generation community. All resources are publicly available at: \url{https://github.com/hpcaitech/Open-Sora}.
\end{abstract}

\section{Introduction}

The field of video generation models has advanced rapidly, with UNet-based architectures demonstrating remarkable performance~\cite{guo2023animatediff,singer2022make,blattmann2023stable}. Following the impressive results showcased by OpenAI’s Sora~\cite{videoworldsimulators2024}, Transformer-based models such as DiT~\cite{peebles2023scalable} have highlighted the potential of scaling diffusion models~\cite{ma2024latte,gupta2025photorealistic,lin2024open}. In this context, we introduce Open-Sora, one of the earliest projects to reproduce Sora, achieving strong results and garnering significant attention.

In the Open-Sora project, we provide comprehensive support for training video generation models, offering the complete suite of necessary code. This includes everything from processing and filtering training data to the training code and model weights, all made available to the community to foster its development. We have successfully reproduced nearly all the techniques mentioned in the Sora report, enabling the generation of videos up to 16 seconds in length, at multiple resolutions up to 720p, with controllable motion dynamics for text-to-video and image-to-video tasks. Figure~\ref{fig:header} showcases examples of videos generated by our model.

\paragraph{Overview} Open-Sora has undergone three major updates, corresponding to versions 1.0 (Mar 2024), 1.1 (April 2024), and 1.2 (June 2024). Each release has an online report in the repository. Unless otherwise specified, Open-Sora specifically refers to Open-Sora 1.2; elsewhere, Open-Sora is used as the default reference. This paper provides a comprehensive overview of the techniques behind these versions and highlights the performance of version 1.2. In Section~\ref{sec:data}, we discuss data composition and preprocessing methods. Section~\ref{sec:model} focuses on the model architecture, and Section~\ref{sec:train} details the training process.

\section{Data}
\label{sec:data}

\subsection{Data Source}

The dataset used is all open-sourced to make the model training fully reproducible. In total, 30M video clips ranging from 2s to 16s are generated, with a total duration of 80k hours. \textbf{Webvid-10M}~\cite{Bain21} contains 10M video-text pairs from the stock footage sites. The videos are low-resolution and have a watermark. \textbf{Panda-70M}~\cite{chen2024panda} is a large-scale dataset with 70M video-caption pairs. We use a 20M high-quality subset for training. \textbf{HD-VG-130M} comprises 130M text-video pairs. The caption is generated by BLIP-2. We find the scene and the text quality are relatively poor. \textbf{MiraData}~\cite{wang2023videofactory} a high-quality dataset with 77k long videos, mainly from games and city exploration. \textbf{Vript}~\cite{yang2024vript} a densely annotated dataset of 400k videos. \textbf{Inter4K}~\cite{stergiou2021adapool} is a dataset containing 1K video clips with 4K resolution.

In addition, we get free-licensed videos from \textbf{Pexels}, \textbf{Pixabay}, and \textbf{Mixkit}. Most videos from these websites are of high quality. We really appreciate these great platforms and the contributors.

The image dataset, which we use to train together with videos, contains around 3M images in total. \textbf{LAION}~\cite{schuhmann2022laion} is a large-scale open dataset, and we use a subset with an aesthetic score larger than 6.5. \textbf{Unsplash-lite}~\cite{unsplash} dataset comprises 25k nature-themed Unsplash photos. This dataset covers a vast range of uses and contexts.

\begin{figure}[!h]
    \begin{center}
        \centerline{\includegraphics[width=0.9\textwidth]{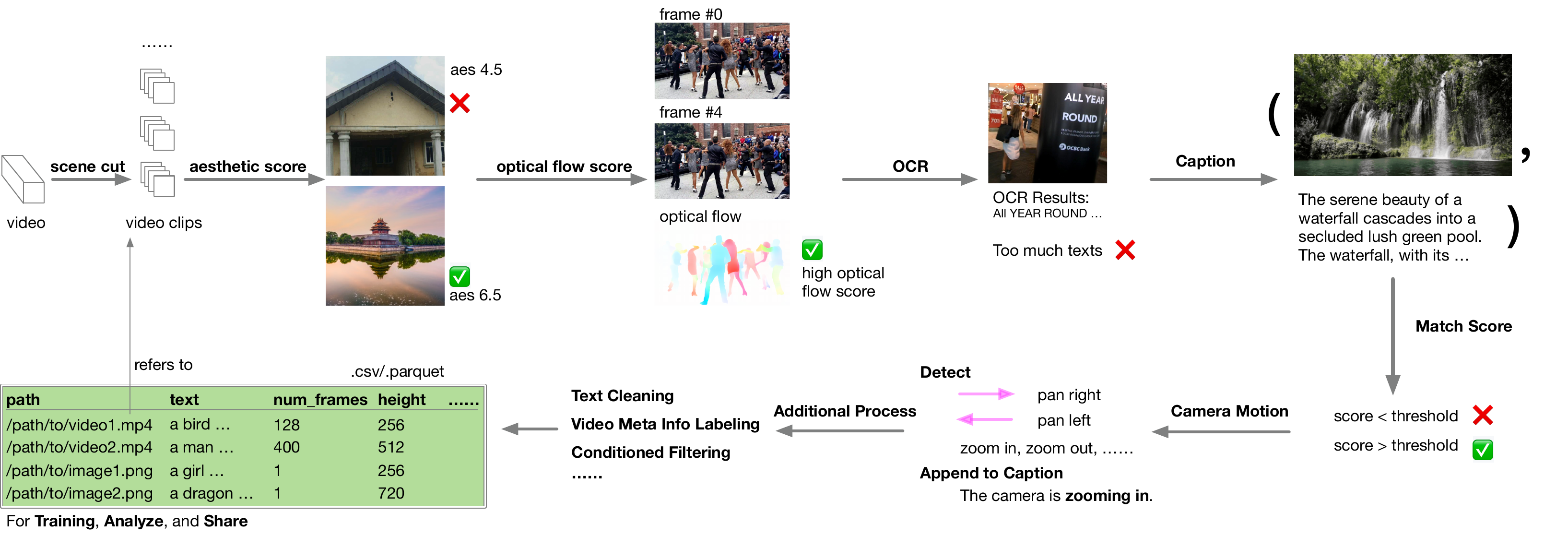}}
        \caption{Open-Sora Data Labeling and Processing Pipeline}
        \label{fig:data_pre}
    \end{center}
\end{figure}

\subsection{Data Pre-processing}

High-quality data is crucial for training good generation models. To this end, we establish a complete pipeline for data processing, which could seamlessly convert raw videos to high-quality video-text pairs. The pipeline is shown in Figure~\ref{fig:data_pre}. To begin the data pre-process pipeline, we use PySceneCut~\cite{pyscenecut} to detect scenes and cut videos into clips.

\begin{figure}[t]
    \begin{center}
        \centerline{\includegraphics[width=0.9\textwidth]{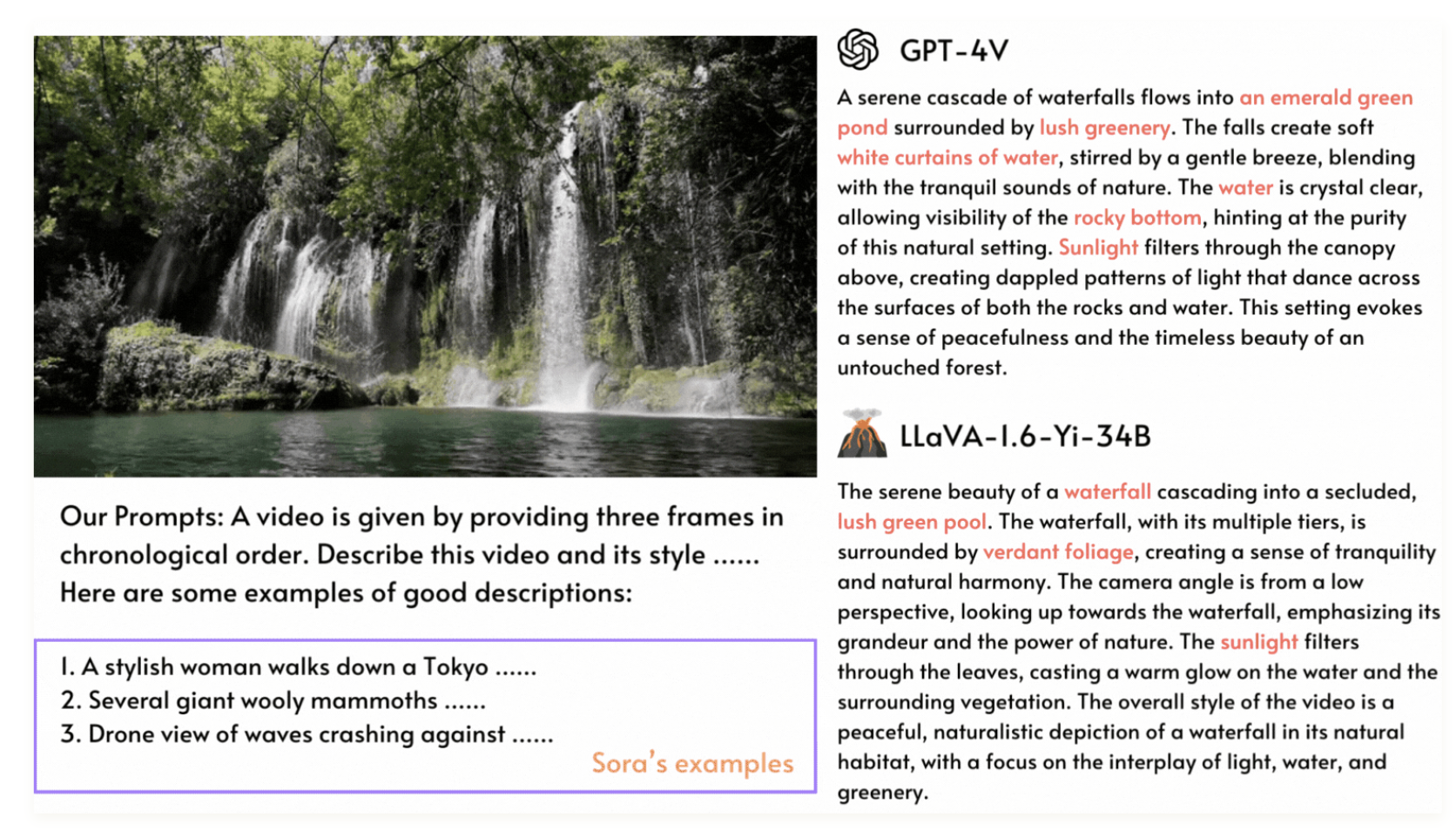}}
        \caption{Open-Sora Video Captioning}
        \label{fig:caption}
    \end{center}
\end{figure}

For high-quality video filtering, we mainly follow the SVD data preprocessing pipeline~\cite{blattmann2023stable}. \textbf{Aesthetic Score} measures the aesthetic preference of frames in the video. We use the scorer from LAION~\cite{schuhmann2022laion} and use the average score among three sampled frames. \textbf{Optical Flow Score} measures the dynamics scale of the video and can be used to filter videos with low motion. We use the UniMatch~\cite{xu2023unifying} model for this task. Some videos are of dense text scenes like news broadcasts and advertisements, which are not desired for training. \textbf{Optical Character Recognition (OCR)} recognizes texts available in the videos, and those with too much text are removed. We use the DBNet++ model implemented by MMOCR~\cite{liao2022real}.

\begin{figure}[b]
    \begin{center}
        \centerline{\includegraphics[width=1.0\textwidth]{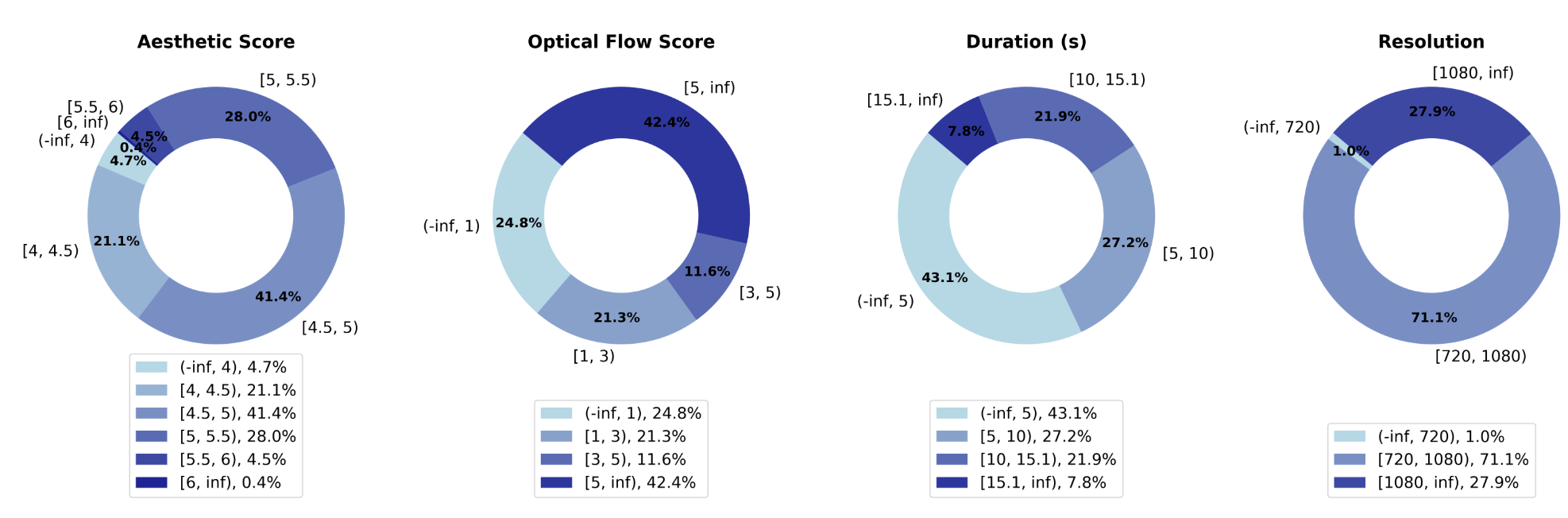}}
        \caption{Distribution of data for last stage training.}
        \label{fig:data_stat}
    \end{center}
\end{figure}

To provide good captions for videos, we utilize GPT-4V and PLLaVA~\cite{xu2024pllava}. The former one provides API, and the latter one is open-source and can be deployed by us. Although with some level of hallucinations, the results are enough to train a text-to-video model. In practice, we use the pretrained PLLaVA 13B model and select 4 frames from each video for captioning with a spatial pooling shape of 2*2. However, the captioning model can hardly produce information about camera movement. Thus, we detect camera motion with optical flow and append the text to the caption.

Some statistics of the video data used in the last stage are shown in Figure~\ref{fig:data_stat}. We present basic statistics of duration and resolution, as well as aesthetic score and optical flow score distribution. We also extract tags for objects and actions from video captions and count their frequencies.

\section{Model Architecture}
\label{sec:model}

Our video generation framework follows Sora's report~\cite{videoworldsimulators2024}. The videos are first compressed by a video compression network, namely a 3d autoencoder. A text encoder encodes texts. Then, a DiT-like transformer handles the video and text latent.

\subsection{3D Autoencoder}

In Open-Sora 1.0 and 1.1, we utilized Stability-AI’s 2D VAE~\cite{rombach2021highresolution} (84M parameters), which compresses video spatially by a factor of 8x8. To reduce the temporal dimension, we downsampled by extracting one frame every three frames. However, this approach resulted in low temporal fluency due to a reduction in generated FPS. To address this limitation, Open-Sora 1.2 introduces a video compression network inspired by OpenAI’s Sora, achieving 4x compression in the temporal dimension. This eliminates the need for frame extraction, enabling video generation at the original FPS.

\begin{figure}[t]
    \begin{center}
        \centerline{\includegraphics[width=0.7\textwidth]{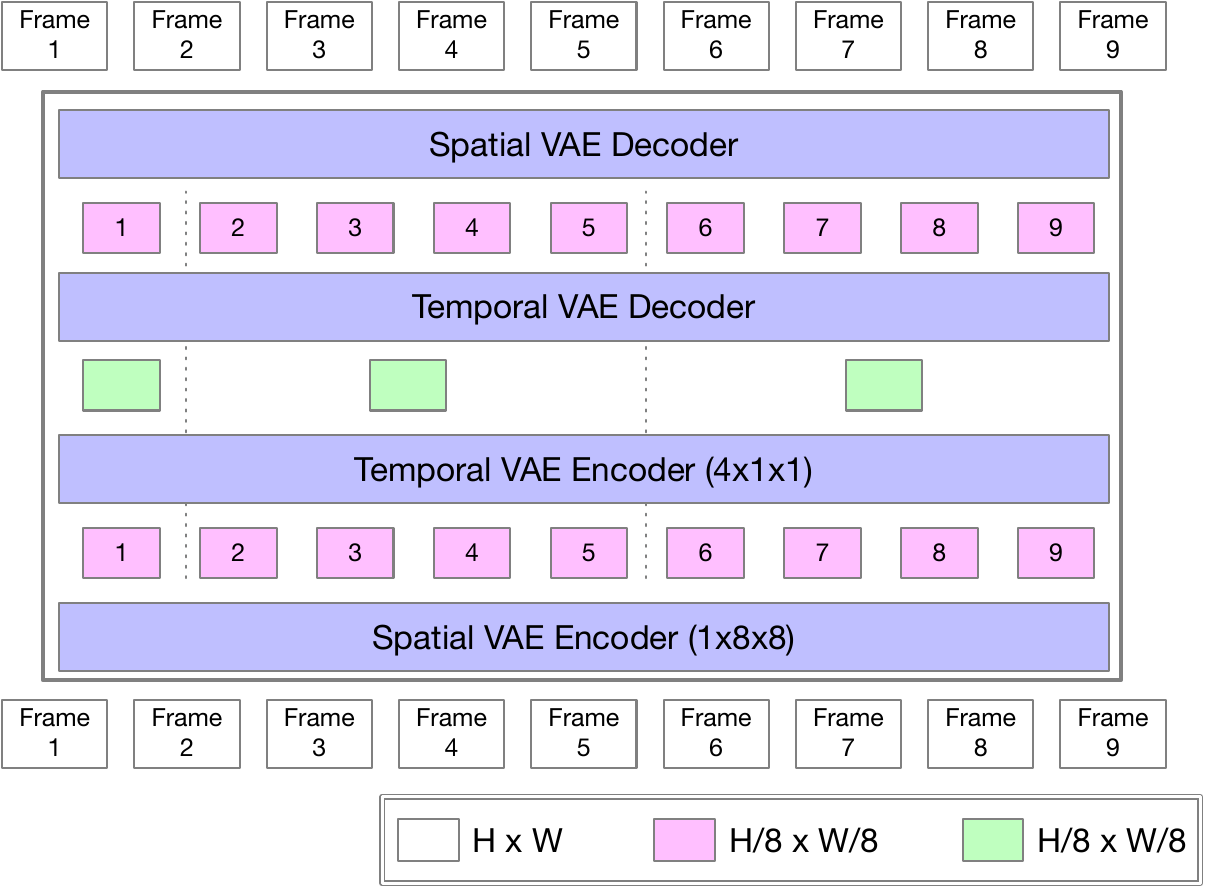}}
        \caption{3D autoencoder by utilizing a pretrained 2D autoencoder.}
        \label{fig:vae}
    \end{center}
\end{figure}

Given the high computational demands of training a 3D VAE, we aimed to leverage the knowledge embedded in the 2D VAE. Post-compression by the 2D VAE, we observed that temporally adjacent features remain highly correlated. Based on this insight, we developed a simple yet effective video compression network that first compresses spatially by 8x8 and subsequently temporally by 4x. The network architecture is illustrated in Figure~\ref{fig:vae}.

We initialized the 2D VAE using SDXL’s pre-trained VAE~\cite{podell2023sdxl}. For the 3D VAE, we adopted the architecture of Magvit-v2’s VAE~\cite{yu2023language}, comprising 300M parameters. Combined with the 2D VAE, the total parameter count of the video compression network is 384M. The 3D VAE was trained for 1.2M steps with a local batch size of 1, using videos from Pexels and Pixabay. The training data primarily consisted of 17-frame clips at a resolution of 256×256. To improve image reconstruction accuracy, causal convolutions were employed within the 3D VAE.

Our training process consists of three stages: \textbf{Stage 1} (0--380k steps) Trained on 8 GPUs with the 2D VAE weights frozen. Objectives included reconstructing 2D VAE-compressed features and applying an identity loss to align features from the 3D VAE with those of the 2D VAE. The identity loss enabled faster convergence and improved initial image reconstruction quality. \textbf{Stage 2} (380k--640k steps) The identity loss was removed, and the 3D VAE was trained to refine its temporal understanding. \textbf{Stage 3} (640k--1.2M steps) Reconstruction of 2D VAE features was found insufficient for further improvement, so the loss was replaced with a direct reconstruction of original videos. This stage utilized 24 GPUs and incorporated mixed video-length training by randomizing video lengths (up to 34 frames) with appropriate zero-padding, improving robustness to varying video durations.

During the first two stages, the dataset comprised 80\% video and 20\% image data. For video training, 17-frame clips were used, while image data was zero-padded to match the input format. However, we found this approach led to blurriness in videos of non-standard lengths. Mixed-length training in Stage 3 effectively resolved this issue.

The stacked VAE architecture requires minimal memory during inference, as inputs are already compressed. For efficiency, input videos are segmented into 17-frame clips. Compared to another open-source 3D VAE~\cite{lin2024open}, our model achieves comparable performance with significantly reduced computational cost.

\begin{table}[t]
    \centering
    \caption{Performance of VAE on validation dataset from pixabay.}
    \begin{tabular}{ccc}
        \toprule
        Model & SSIM↑ & PSNR↑ \\
        \midrule
        Open-Sora-Plan 1.1 & 0.882 & 29.890 \\
        Open-Sora 1.2 & 0.880 & 30.590 \\
        \bottomrule
    \end{tabular}
\label{tab:diff}
\end{table}

\subsection{Architecture}

\begin{wrapfigure}{l}{6.5cm}%
    \begin{center}
        \centerline{\includegraphics[width=0.35\textwidth]{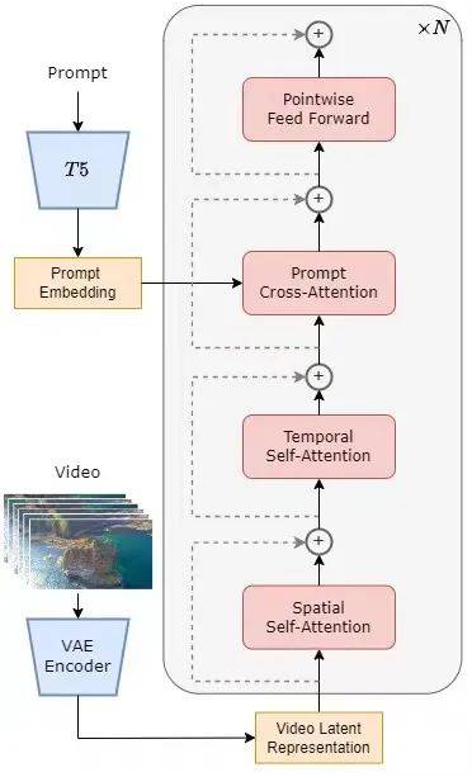}}
        \caption{Open-Sora diffusion transformer architecture.}
        \label{fig:arch}
    \end{center}
\end{wrapfigure}%

Our model architecture builds upon PixArt~\cite{chen2023pixart}, an image diffusion transformer, where text is encoded using a T5 text encoder~\cite{raffel2020exploring}, and cross-attention is applied between video and text latents. To enable efficient video generation, we employ a spatial-temporal attention mechanism, Spatial-Temporal Diffusion Transformer (STDiT), inspired by Latte~\cite{ma2024latte}, replacing full attention on all tokens. Specifically, spatial self-attention is applied within each frame, while temporal attention is applied across frames at the same spatial location.

To focus on video generation, we designed the model to build upon a robust pre-trained image generation model. The model is initialized with PixArt-$\alpha$, a T5-conditioned DiT structure optimized for high-quality and efficient image generation. The projection layers for the newly introduced temporal attention are initialized to zero, preserving the model's original image generation capabilities at the start of training. The inclusion of temporal attention doubles the parameter count from 580M to 1.1B.

Several modifications were introduced to enhance training stability and performance. Following best practices in large language models, we replaced sinusoidal positional encoding with rotary positional embeddings (RoPE)~\cite{su2023enhanced} for temporal attention, as this aligns better with the sequence prediction nature of video generation tasks. Additionally, inspired by SD3~\cite{esser2024scaling}, we applied QK-normalization to all attention mechanisms to improve training stability, particularly under half-precision training. We used a small epsilon ($1 \times 10^{-15}$) for QK-normalization, which proved effective in avoiding training spikes and ensuring smoother optimization~\cite{molybog2023theory}. These enhancements, combined with the spatial-temporal attention mechanism, enable our model to transition seamlessly from high-quality image generation to efficient video generation while maintaining stability and performance throughout the training process.

\subsection{Conditioning}

\begin{figure}[t]
    \begin{center}
        \centerline{\includegraphics[width=0.8\textwidth]{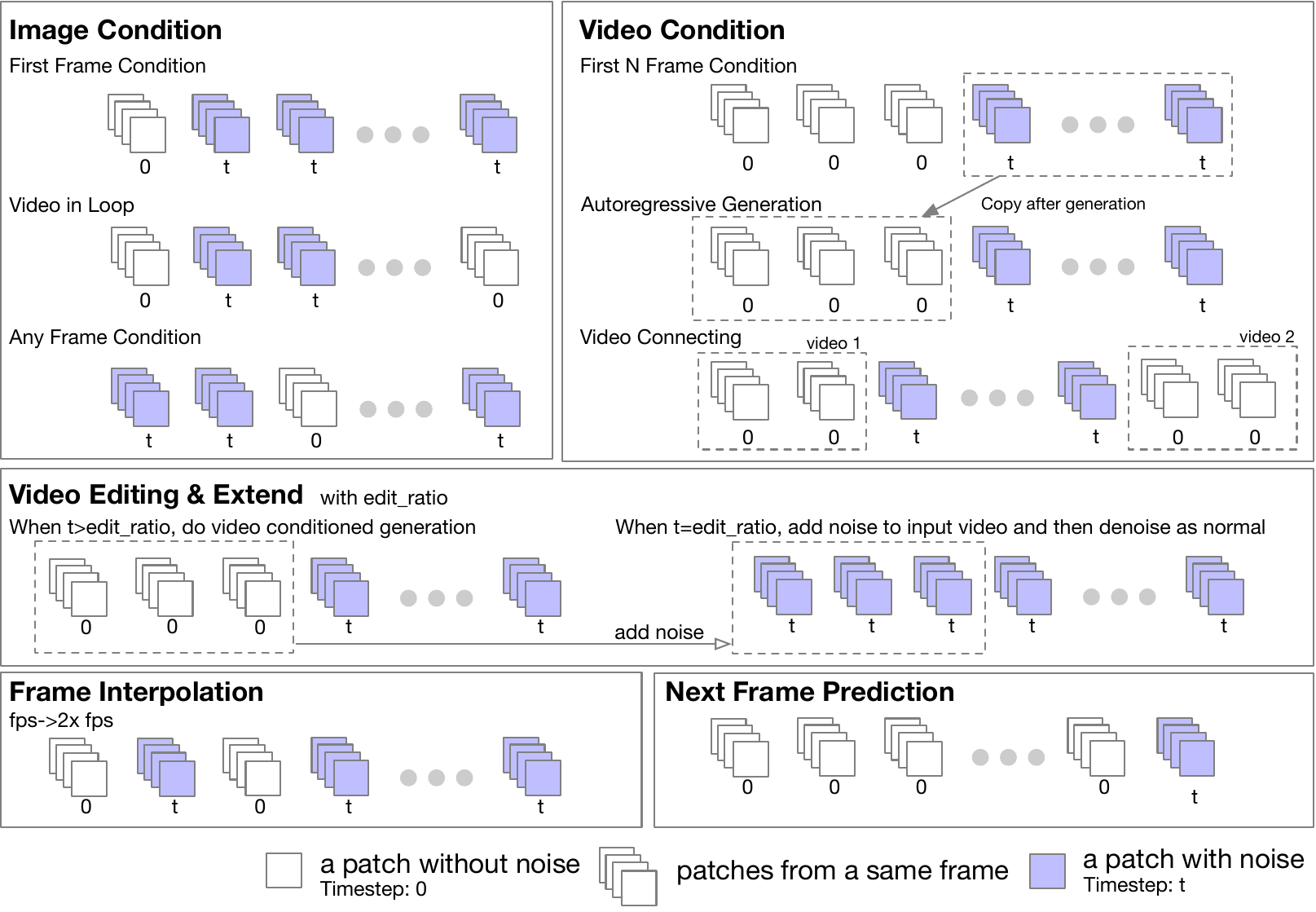}}
        \caption{A general framework for image and video to video generation.}
        \label{fig:mask}
    \end{center}
\end{figure}

While text-to-video generation is highly versatile, certain applications require greater control. Transformers are inherently adaptable and can be extended to support tasks such as image-to-image and video-to-video generation. To enable such conditioning, we propose a masking strategy for image and video inputs, as illustrated in Figure~\ref{fig:mask}.

In this strategy, frames designated for conditioning are unmasked. During the forward pass, these unmasked frames are assigned a timestep of 0, while other frames retain their diffusion timesteps. However, directly applying this strategy to a pre-trained model often yields suboptimal results, as the diffusion model has not been trained to manage mixed timesteps within a single sample.

Inspired by UL2~\cite{tay2022ul2}, we address this issue by introducing a random masking strategy during training. Specifically, frames are randomly unmasked with patterns such as the first frame, the first k frames, the last frame, the last k frames, a combination of the first and last k frames, or entirely random frames. Using Open-Sora 1.0 as a baseline, we experimented with masking applied to 50\% of training samples and observed that the model effectively learned image and video conditioning capabilities after 10k steps, with minimal impact on text-to-video performance. A lower masking probability (e.g., 30\%) resulted in reduced conditioning effectiveness. Consequently, we pre-trained the model from scratch with this masking strategy.

To further enhance the model’s control capabilities, we appended scores to captions, using them as additional conditioning inputs. These scores include aesthetic scores, motion scores, and camera motion descriptors. For instance, a caption for a video with an aesthetic score of 5.5, a motion score of 10, and detected camera motion of “pan left” would be formatted as: \textit{[Original Caption] aesthetic score: 5.5, motion score: 10, camera motion: pan left.} During inference, these scores can also be adjusted to influence video generation. For camera motion conditioning, we manually labeled 13,000 high-confidence clips.

This approach provides the model with a nuanced understanding of conditioning inputs, improving its ability to generate high-quality, context-aware videos across a variety of tasks.

\section{Training Strategy}
\label{sec:train}

Training video generation models demands substantial computational resources. To optimize efficiency, we adapt a pre-trained image generation model to a video generation model and employ a multi-stage training strategy. Specifically, we utilize flow matching~\cite{lipman2022flow} with a learning rate of $5 \times 10^{-5}$. The entire training process spans 68k steps and requires approximately 35,000 H100 GPU hours. These approachs significantly reduces the training cost while achieving high-quality video generation performance.

\subsection{Multi-resolution and Multi-aspect-ratio}

As highlighted in Sora’s report, training with the original resolution, aspect ratio, and length of videos improves sampling flexibility and enhances framing and composition. To achieve this goal, we evaluated three approaches:

\begin{itemize}
    \item NaViT~\cite{dehghani2024patch}: This method supports dynamic sizes within the same batch through masking with minimal efficiency loss. However, its implementation is complex and may not fully leverage optimized kernels like Flash Attention~\cite{dao2022flashattention}.
    \item Padding (FiT~\cite{lu2024fit}): This approach supports dynamic sizes within the same batch by padding smaller resolutions to match the largest one. While simple, padding results in inefficient memory usage for varying resolutions.
    \item Bucket (SDXL~\cite{podell2023sdxl}, PixArt~\cite{chen2023pixart}): This method supports dynamic sizes across different batches by grouping samples into pre-defined “buckets.” Within each batch, the resolution, frame count, and aspect ratio are fixed. Bucketing avoids the complexities of masking or padding and benefits from optimized operations on uniform-sized inputs. However, it limits flexibility to a fixed set of sizes.
\end{itemize}

For simplicity and efficiency, we adopt the bucket-based approach. We pre-define a set of fixed resolutions, aspect ratios, and frame lengths and allocate samples to buckets accordingly. Each bucket is defined as a triplet of (resolution, number of frames, aspect ratio) that covers most common video formats. Before each training epoch, the dataset is shuffled, and samples are assigned to the largest bucket that fits their resolution and frame length.

To further optimize computational resources, we introduce two additional attributes for each bucket: probability of keeping in the bucket and batch size. According to the probability, high-resolution videos are downsampled to a lower resolution, effectively reducing computational cost. Batch sizes are adjusted per bucket to balance GPU load, ensuring efficient resource utilization. By fine-tuning these parameters, we achieve a balanced distribution of samples across buckets and improve overall training efficiency while maintaining high-quality video generation.

This bucket-based strategy provides a practical trade-off between implementation simplicity and computational efficiency, enabling flexible training with diverse video resolutions and aspect ratios.

\subsection{Model Adaptation}

We begin with the PixArt-$\Sigma$ 2K checkpoint~\cite{chen2025pixart}, a model trained with DDPM~\cite{ho2020denoising} and SDXL VAE at much higher resolutions. The model is efficiently adapted to video generation tasks by finetuning on a smaller dataset. The adaptation process comprises several sequential stages, all conducted on 8 H100 GPUs.

\begin{enumerate}
    \item Multi-resolution image generation: Training the model to handle resolutions from 144p to 2K over 20k steps.
    \item QK-normalization integration: Adding QK-norm for stability, 18k training steps.
    \item Transition to rectified flow: Shifting from discrete-time DDPM to continuous-time rectified flow, with 10k training steps.
    \item Enhanced rectified flow training: Incorporating logit-norm sampling and resolution-aware timestep sampling, training for 33k steps.
    \item Smaller AdamW epsilon: Following SD3, using a reduced epsilon ($1 \times 10^{-15}$) with QK-norm for stability, trained for 8k steps.
    \item New VAE and FPS conditioning: Replacing the original VAE with Open-Sora’s, adding FPS conditioning to timestep conditioning, and training for 25k steps. Normalizing each channel proved crucial for rectified flow training.
    \item Temporal attention blocks: Adding zero-initialized temporal attention blocks, initially trained on images for 3k steps.
    \item Mask strategy for temporal blocks: Focusing temporal attention blocks exclusively on videos using a masking strategy, trained for 38k steps.
\end{enumerate}

After completing this adaptation, the model retains its ability to generate high-quality images while gaining multiple benefits for video generation.

\begin{enumerate}
    \item Accelerated training and inference: Rectified flow reduces the number of sampling steps from 100 to 30 for videos, significantly decreasing inference time.
    \item Enhanced stability: QK-norm enables more aggressive optimization, improving training efficiency.
    \item Efficient temporal compression: The new VAE compresses the temporal dimension by a factor of 4, reducing the computational cost.
    \item Resolution flexibility: The model can generate videos at multiple resolutions, from 144p to 2K, supporting diverse use cases.
\end{enumerate}

This comprehensive adaptation not only enhances the model’s video generation capabilities but also ensures efficient and scalable training, setting a new standard for open-source diffusion-based video generation.

\subsection{Multi-stage Training}

\begin{figure}[t]
    \begin{center}
        \centerline{\includegraphics[width=0.8\textwidth]{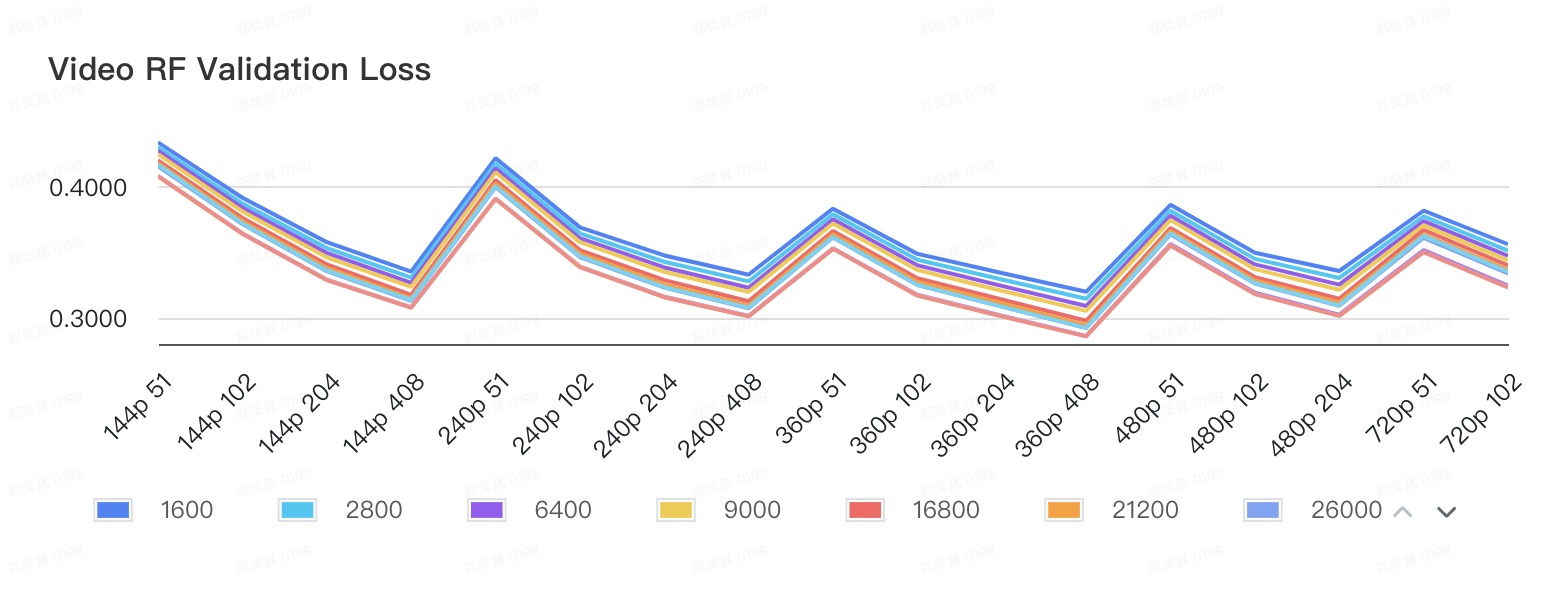}}
        \caption{Validation loss for varying lengths and different resolutions.}
        \label{fig:val_loss}
    \end{center}
\end{figure}

\begin{figure}[b]
    \centering
    \begin{minipage}[b]{0.5\textwidth}
        \centering
        \includegraphics[width=\textwidth]{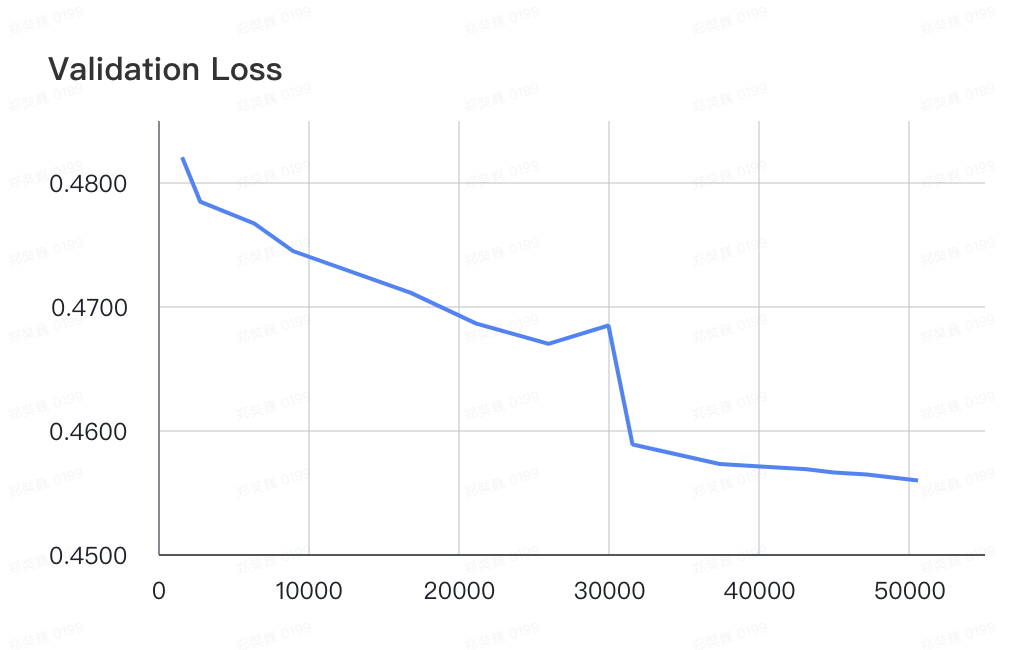}
        \label{fig:val_loss_2}
    \end{minipage}
    \hfill
    \begin{minipage}[b]{0.45\textwidth}
        \centering
        \includegraphics[width=\textwidth]{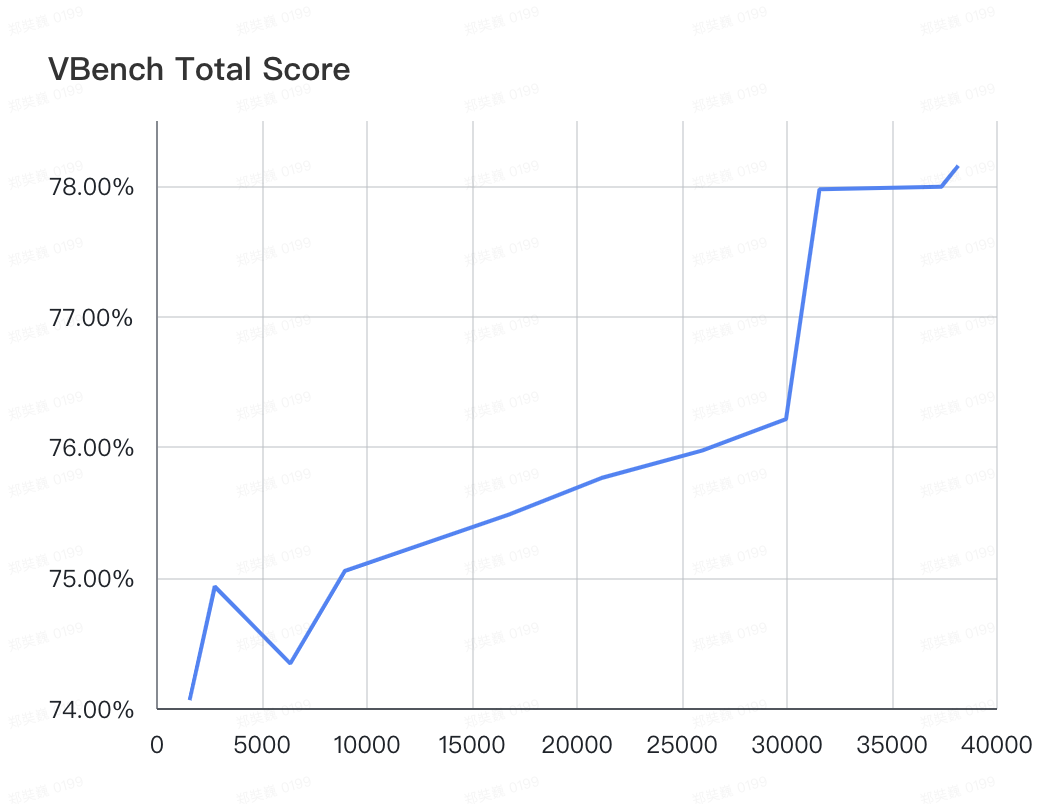}
        \label{fig:vbench}
    \end{minipage}
    \caption{Validation loss and VBench score during training.}
    \label{fig:val_vbench}
\end{figure}

To optimize performance within a limited computational budget, we carefully organized the training data by quality and split the training process into three stages. The model was trained on a 12×8 GPU setup over approximately two weeks, completing around 70k steps.

In the first stage, the model was trained on the Webvid-10M dataset (40k hours of video) for 30k steps (2 epochs). This dataset primarily contains videos with resolutions below 360p and watermarks, making it ideal for initial training. We focused on videos with resolutions of 240p and 360p, spanning lengths of 2 to 16 seconds. The original dataset captions were used for training.

In the second stage, we trained the model on the Panda-70M dataset. Given the variable quality of this dataset, we used the official 30M subset, filtered to retain only videos with aesthetic scores above 4.5, resulting in a 20M subset (41k hours). Training primarily focused on resolutions of 360p and 480p for 23k steps, equivalent to 0.5 epochs. Although this training stage was not fully completed, it provided sufficient improvements to prepare the model for broader use.

The final stage involved a curated collection of approximately 2M high-quality video clips from diverse sources, totaling 5k hours. While videos from MiraData and Vript included captions generated by GPT, other sources were captioned using PLLaVA. This stage focused on higher resolutions (720p and 1080p) to enhance the model’s capability to handle larger resolutions. A 25\% masking ratio was applied during training, which spanned 15k steps (around 2 epochs).

For validation, we sampled 1k videos from Pixabay to evaluate the model’s performance. The evaluation loss was calculated for images and videos of varying lengths (2s, 4s, 8s, 16s) across different resolutions (144p, 240p, 360p, 480p, 720p). Losses were averaged over 10 equidistant timesteps for each configuration.

We also tracked VBench scores during training. VBench~\cite{huang2024vbench} is an automated benchmark for evaluating short video generation. Scores were computed using 240p 2-second videos, providing additional validation of the model’s progress. Both evaluation loss and VBench scores confirm consistent improvements in the model’s performance throughout the training process. The VBench score and validation loss during training are shown in Figure~\ref{fig:val_vbench}.

More samples of video generation are available at \url{https://hpcaitech.github.io/Open-Sora/}. Table~\ref{tab:benchmark} presents the VBench scores of various models, demonstrating that Open-Sora achieves state-of-the-art performance in video generation among open-source models. 

\begin{table}[t]
    \centering
    \caption{VBench score comparison of different open-source models.}
    \begin{tabular}{lccc}
        \toprule
        Model & Total Score (\%) & Quality Score (\%) & Semantic Score (\%) \\
        \midrule
        CogVideo~\cite{hong2022cogvideo} & 67.01 & 72.06 & 46.83 \\
        Latte~\cite{ma2024latte} & 77.29 & 79.72 & 67.58 \\
        LaVie~\cite{wang2024lavie} & 77.08 & 78.78 & 70.31 \\
        Show-1~\cite{zhang2024show} & 78.93 & 80.42 & 72.98 \\
        OpenSoraPlan V1.1~\cite{lin2024open} & 78.00 & 80.91 & 66.38 \\
        OpenSoraPlan V1.2~\cite{lin2024open} & 75.98 & \textbf{81.51} & 53.88 \\
        OpenSoraPlan V1.3~\cite{lin2024open} & 77.23 & 80.14 & 65.62 \\
        \midrule
        Open-Sora 1.0 & 75.91 & 78.82 & 64.28 \\
        Open-Sora 1.1 & 75.66 & 77.74 & 67.36 \\
        Open-Sora 1.2 & \textbf{79.76} & 81.35 & \textbf{73.39} \\
        \bottomrule
    \end{tabular}
    \vspace{0.5cm}
\label{tab:benchmark}
\end{table}

\section{Conclusion}
Open-Sora represents a significant step forward in open-source video generation, providing a comprehensive framework that includes data processing, training code, and model weights. By successfully reproducing key techniques from the Sora report and enabling the generation of high-quality videos up to 16 seconds long, with resolutions up to 720p and controllable motion dynamics, Open-Sora democratizes access to advanced video generation technology. This initiative not only fosters community collaboration but also sets a foundation for future advancements in the field.

\newpage
\printbibliography

\end{document}